\documentclass[conference,a4paper]{IEEEtran}
\usepackage{listings}
\usepackage[style=base]{caption} 
\usepackage{float}

\ifCLASSINFOpdf
  \usepackage[pdftex]{graphicx}
  \DeclareGraphicsExtensions{.pdf,.jpeg,.jpg,.png}
\else
  \usepackage[dvips]{graphicx}
  \DeclareGraphicsExtensions{.eps}
\fi
\usepackage[cmex10]{amsmath}
\ifCLASSOPTIONcompsoc
  \usepackage[caption=false,font=normalsize,labelfont=sf,textfont=sf]{subfig}
\else
  \usepackage[caption=false,font=footnotesize]{subfig}
\fi
\begin{document}
%
\title{Automatic Ground Truths: Projected Image Annotations for Omnidirectional Vision}

\author{\IEEEauthorblockN{Victor Stamatescu\IEEEauthorrefmark{1},
Peter Barsznica\IEEEauthorrefmark{1},
Manjung Kim \IEEEauthorrefmark{1},
Kin K. Liu\IEEEauthorrefmark{1},
Mark McKenzie\IEEEauthorrefmark{2},
\\Will Meakin\IEEEauthorrefmark{1},
Gwilyn Saunders\IEEEauthorrefmark{1},
Sebastien C. Wong\IEEEauthorrefmark{2} and
Russell S. A. Brinkworth\IEEEauthorrefmark{1}}
\IEEEauthorblockA{\IEEEauthorrefmark{1}University of South Australia, Mawson Lakes, SA, Australia
\\Email: Victor.Stamatescu@unisa.edu.au}
\IEEEauthorblockA{\IEEEauthorrefmark{2}Defence Science and Technology Group, Edinburgh, SA, Australia}
}



\maketitle

\begin{abstract}
We present a novel data set made up of omnidirectional video of multiple objects whose centroid positions are annotated automatically. Omnidirectional vision is an active field of research focused on the use of spherical imagery in video analysis and scene understanding, involving tasks such as object detection, tracking and recognition. Our goal is to provide a large and consistently annotated video data set that can be used to train and evaluate new algorithms for these tasks. Here we describe the experimental setup and software environment used to capture and map the 3D ground truth positions of multiple objects into the image. Furthermore, we estimate the expected systematic error on the mapped positions. In addition to final data products, we release publicly the software tools and raw data necessary to re-calibrate the camera and/or redo this mapping. The software also provides a simple framework for comparing the results of standard image annotation tools or visual tracking systems against our mapped ground truth annotations.
\end{abstract}


%
\IEEEpeerreviewmaketitle

\section{Introduction}

This paper presents a new data set for omnidirectional vision together with a novel approach to ground truth image annotation. Our work is motivated by the emerging use of omnidirectional imagery in vision applications, where new dedicated image processing techniques are yet to reach levels of maturity and benchmarking generally found elsewhere in computer vision (e.g. see~\cite{demonceaux2009omnidirectional} and references therein). In order to facilitate the development and evaluation of new algorithms in this area, we have built a large data set with consistent ground truth image annotations.

\begin{figure}[!t]
\centering
\includegraphics[width=2.35in]{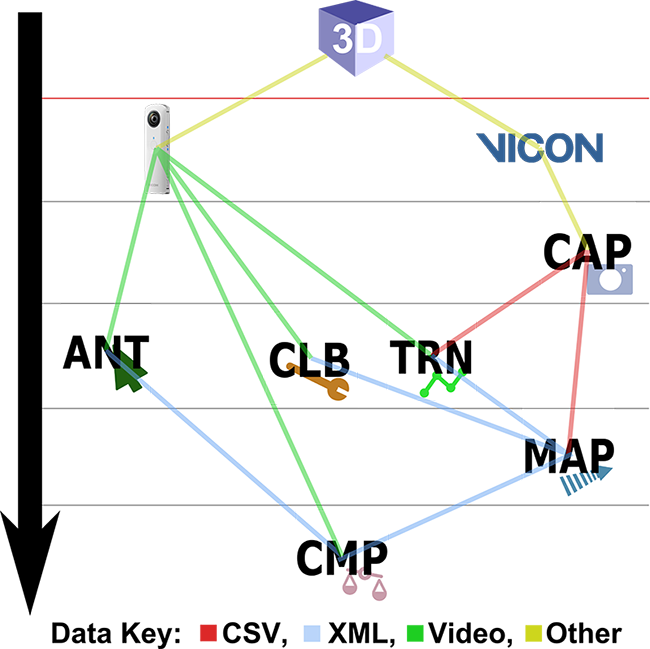}
\caption{Overview of the UniSA omnidirectional data set acquisition and automatic annotation. RICOH THETA m15 image (2D) and VICON positional (3D) centroid data were captured (CAP) in the lab. Point correspondences from chessboard calibration images (CLB) and training data from a VICON calibration wand (TRN) were used to compute the intrinsic camera parameters and its pose, respectively. This enabled the projection (MAP) of 3D target object positions measured by the VICON system to 2D image centroids. These annotations may then be used to evaluate (CMP) the performance of a human annotator or visual tracking system (ANT).}
\label{fig_dataflow}
\end{figure}

Current benchmarks for multi-object visual tracking and recognition (e.g.~\cite{MOTChallenge2015,MOT16,wen2015detrac,kasturi2014performance}) have employed human annotators together with semi-automatic image annotation tools to supply ground truth data, which introduces unknown bias into the performance evaluation of machine vision systems~\cite{milan2013challenges}. Unlike this previous work, our data set provides ground truth image annotations that have been automatically mapped from 3D object positions measured by a VICON motion capture system, using a process illustrated in Figure~\ref{fig_dataflow}. This approach avoids ad hoc error prone human annotation and allows the systematic error on the ground truth image annotations to be estimated.

Omnidirectional vision systems have found application in autonomous driving technology~\cite{amuse2013gt}, traffic surveillance~\cite{wang2015real} and threat warning~\cite{brusgard1999distributed}, where large fields of view may be coupled with machine vision to provide complete situational awareness of surroundings. Such systems can consist of multiple sensors whose images are fused together~\cite{peng2015low}. Alternatively, they may employ dioptric (fisheye lenses) or catadioptric (hyperbolic/parabolic/spherical mirrors) optics to provide a spherical field of view. Omnidirectional cameras enable object tracking and recognition over a wider area and longer length of time than can be achieved using standard perspective projection cameras, but this typically comes at the expense of image resolution and additional distortion. The UniSA omnidirectional data set aims to promote research in this area by enabling the training and evaluation of omnidirectional vision algorithms for the detection, tracking and fine-grained recognition of multiple generic objects.

In this paper we describe the UniSA omnidirectional data set, which is made available publicly through our website\footnote{http://www.cls-lab.org/data/unisa-omnidata}. The work presented consists of the following key contributions:
\begin{itemize}
\item A novel approach to ground truth image annotation for visual tracking and object recognition data sets that is automated and avoids the need for human annotators.
\item A set of $43$ omnidirectional videos of moving and stationary target objects in  a variety of scenarios, which include occlusions, lighting changes, clutter and fog.
\item A corresponding set of ground truth centroid positions for each object unique identifier (ID), which have been mapped into the image from precisely measured 3D world coordinates.
\item The publicly available software environment, which provides a suite of tools for data capture, camera calibration, ground truth mapping and comparison with results obtained by human annotators or visual tracking systems.
\item Raw calibration and 3D position data to which the software tools may be applied to re-calibrate the camera and/or re-map the ground truth annotations, respectively.
\end{itemize}

The rest of this paper is organized as follows.
Section~\ref{relatedwork} presents a review of key recent developments in omnidirectional vision research and relevant benchmark data sets. Section~\ref{unisadataset} describes the experimental setup, methodology and the resulting data products of the UniSA omnidirectional data set. Section~\ref{performance} outlines how these data may be used to evaluate omnidirectional vision systems. Finally, section~\ref{conclusion} concludes the paper, outlining directions of future work to advance the data set.

\section{Related Work} \label{relatedwork}

\begin{figure}[!t]
\centering
\includegraphics[width=1.6in,trim={0.cm 0.cm 0.cm 0.7cm},clip]{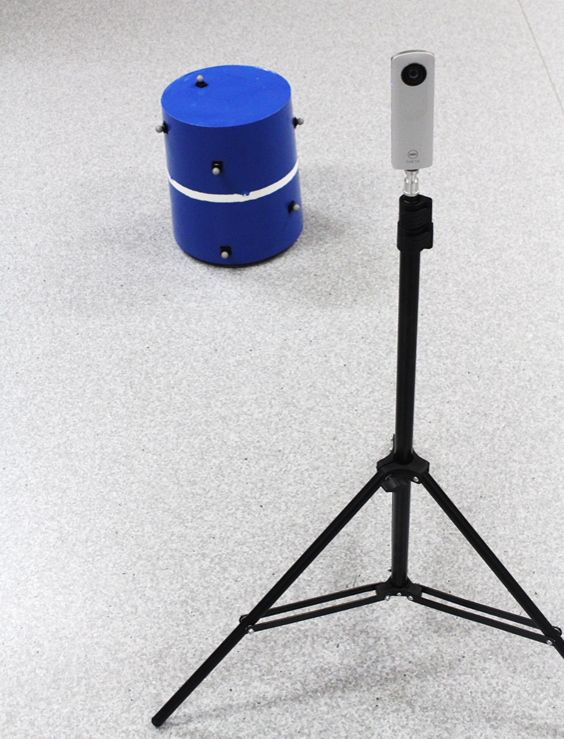}
\caption{RICOH THETA m15 omnidirectional camera mounted on a tripod and a target object in the background. Each target consists of a MechBot covered by a container whose surface is sparsely covered with {9 mm} diameter markers, which define a unique constellation that can be tracked by the lab's VICON system.}
\label{fig_tripod}
\end{figure}

Benchmarks for multi-object visual tracking typically exercise only a specific type of object, such as pedestrians in the MOT Challenge~\cite{MOTChallenge2015,MOT16} or vehicles in UA--DETRAC~\cite{wen2015detrac}, to which highly tuned object detectors can be applied. By contrast, the DARPA Neovision2 benchmark~\cite{kasturi2014performance} captured video of multiple object types to assist the development of detection and recognition algorithms. A subset of its ground truth data has also been extended to include unique object IDs for tracking evaluation~\cite{chakraborty2017data}. In addition, the KITTI~\cite{Geiger2012CVPR} and {ImageNet VID}~\cite{russakovsky2015imagenet} benchmarks are designed for multi-class, multi-object detection and tracking. Unlike the aforementioned data sets, ours captures both stationary and moving targets that are generic objects of rigid shape. Moreover, our videos were recorded using an omnidirectional camera, as shown in Figure~\ref{fig_tripod}.

Depending on the specific geometry of the omnidirectional camera optics, the resultant projection can lead to severe image distortion. This presents a challenge for standard image processing techniques and a number of dedicated solutions have been proposed in the context of structure from motion~\cite{demonceaux2009omnidirectional,taira2015robust}, 3D reconstruction~\cite{ma20153d}, visual odometry~\cite{hadj2008spherical}, object detection~\cite{cinaroglu2016direct} and tracking~\cite{salazar2009visual,wang2015real} using omnidirectional imagery.

Research in these areas has also led to the release of several task--specific omnidirectional data sets focusing on structure from motion~\cite{bastanlar2012multi,zamir2014image}, visual odometry~\cite{amuse2013gt,zhang2016benefit}, human detection~\cite{cinaroglu2014direct} and tracking~\cite{demiroz2012feature,PIROPOdatabase}, and vehicle detection~\cite{karaimer2014car} and recognition~\cite{karaimer2015detection,barics2016classification}.
We note that the ground truth image annotations provided with each of these data sets are reliant on human input, with the exception of AMUSE~\cite{amuse2013gt}. Similar to our approach, the authors provide raw sensor recordings of the environment, however, unlike our data set, they leave it to the end user to determine the ground truths.

The KITTI benchmark~\cite{Geiger2012CVPR} consists of video captured using two stereo camera rigs (grayscale and color) that are combined with localization sensor data for the recording platform and 3D point cloud data from a Velodyne laser scanner. The aim is to evaluate computer vision tasks required in autonomous driving, which include stereo estimation, optical flow, visual odometry and 3D object tracking. While not omnidirectional, this data set is conceptually similar to ours in terms of its ground truth data, which is mapped into the image~\cite{geiger2013vision}. We note, however, that their ground truth 3D bounding box tracklets are still assigned to dynamic objects by human annotators, and this is a key point of difference to our approach, which removes humans from the ground truth annotation process.



\section{UniSA Omnidirectional Data Set} \label{unisadataset}

The UniSA omnidirectional data set consists of $43$ spherical videos and corresponding ground truth object positions measured in 3D world coordinates by a VICON system. These raw data were captured over four sessions in the UniSA Mechatronics Laboratory. This section describes the experimental setup and software tools used to implement the process illustrated by Figure~\ref{fig_dataflow}, together with the resulting data set.

\subsection{Experimental Setup}

Videos were recorded with a RICOH THETA m15 spherical camera whose dual fisheye lens system provides a spherical ($4\pi$ steradian) field of view. The camera remained stationary during each recording session, being mounted on a tripod as shown in Figure~\ref{fig_tripod} and having recordings triggered via a smart phone application. Raw videos were recorded at $15$ frames per second in MOV format with  MPEG-4 AVC/H.264 compression. Each HD ($1920 \times 1080$) resolution image frame contains two fisheye images that correspond to opposite hemispheres, which are treated separately as side-by-side $960 \times 1080$ pixel sized images~\cite{RICOHTHETASpecs}.

\begin{figure}[!t]
\centering
\includegraphics[width=3.1in,trim={1cm 19.5cm 1cm 0},clip]{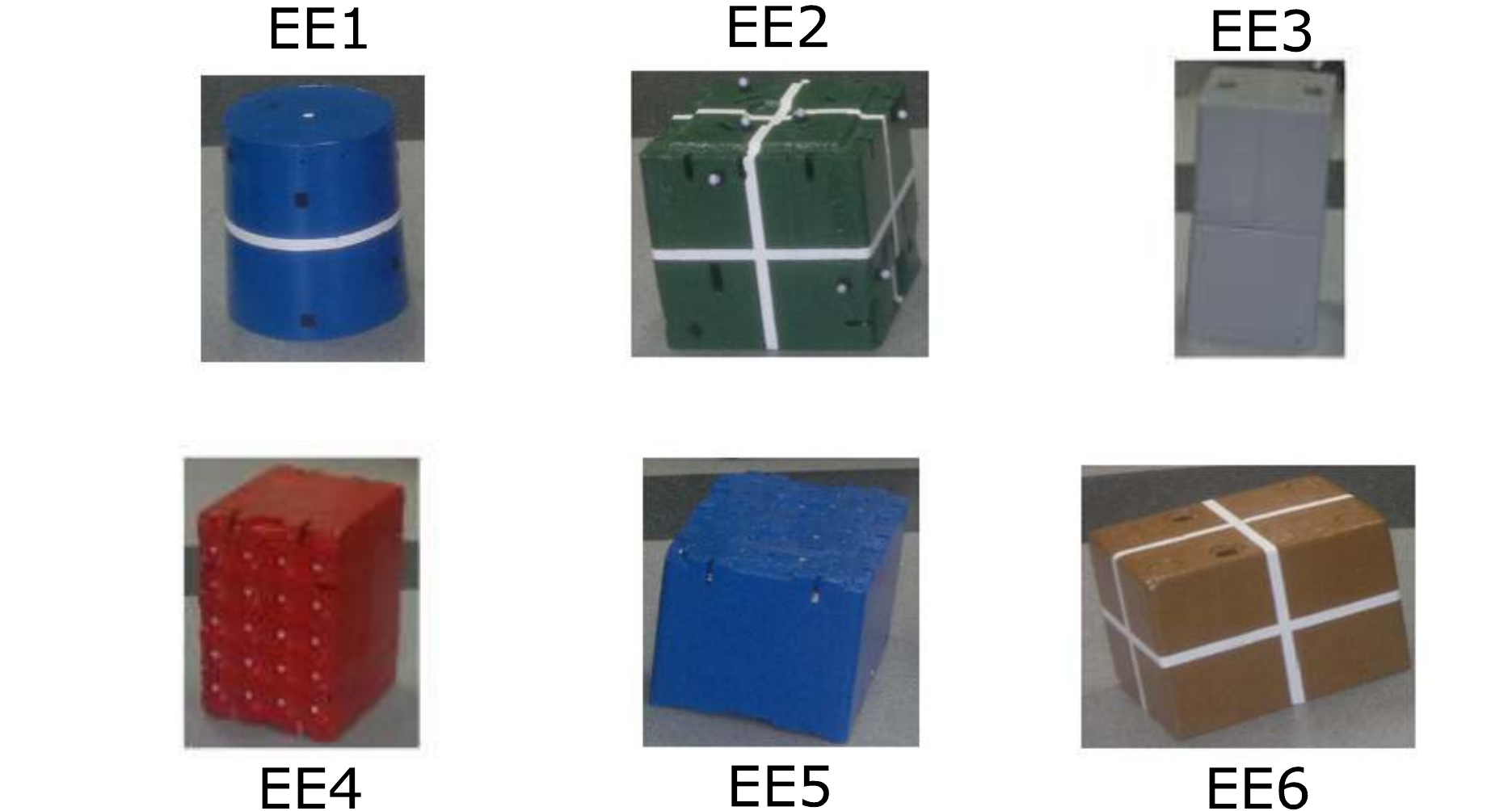}
\caption{The six target objects present in the UniSA omnidirectional data set: blue tub EE1, green box EE2, tall gray box EE3, tall large red box EE4, long blue box EE5 and small brown box EE6.}
\label{fig_boxes}
\end{figure}

\begin{figure}[!t]
\centering
\includegraphics[width=1.9in]{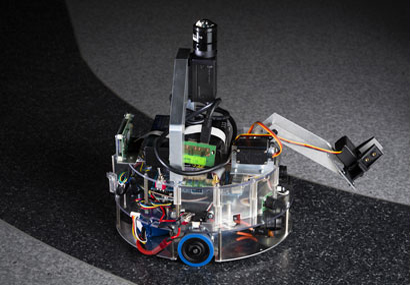}
\caption{Mechatronic Robots (MechBots) are used to move each target and are controlled using PlayStation 3 controllers.}
\label{fig_mechbot}
\end{figure}

\begin{figure}[!t]
\centering
\includegraphics[width=2.6in,trim={0.cm 0.8cm 0.cm 0.8cm},clip]{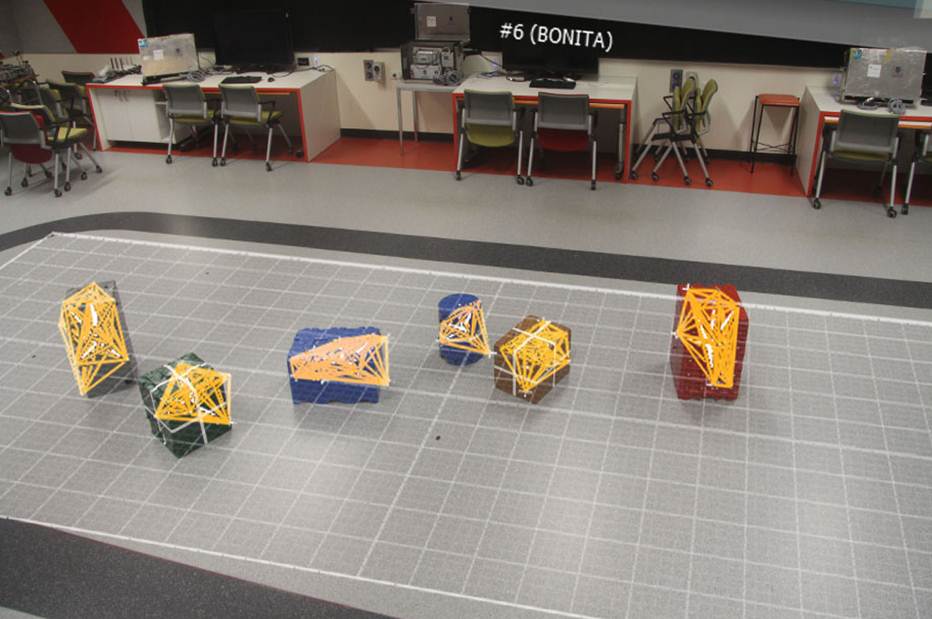}
\caption{VICON target objects (yellow constellations defined by markers) as displayed using the VICON motion capture system software and superimposed on an image of the scene.}
\label{fig_bonita}
\end{figure}

Up to six targets were captured as part of the data set and these are shown in Figure~\ref{fig_boxes}, with their dimensions listed in Table~\ref{targets}. Remote controlled round ground-based robots of diameter $250$ mm with two drive wheels and two jockey wheels (MechBots), such as that seen in Figure~\ref{fig_mechbot}, were placed under each container to enable target motion. The position of each object was tracked with sub-mm accuracy by a VICON system recording at around $40$ Hz. The 3D tracking is based on the unique constellation of markers on each object that is observed by eight Bonita B10 cameras, and this is illustrated in Figure~\ref{fig_bonita} for one of the camera views.

\begin{table}
\centering
\begin{tabular}{l| l}
 & dimensions \\\hline
 EE1 & diameter $= 85$ mm height $= 315$ mm \\
 EE2 & $330\times330\times330$ mm \\
 EE3 & $280\times280\times555$ mm \\
 EE4 & $380\times335\times490$ mm \\
 EE5 & $330\times495\times280$ mm \\
 EE6 & $290\times385\times270$ mm 
\end{tabular}
\caption{\label{targets} Target object dimensions}
\end{table}

Our Python software environment includes the VICON capture tool (CAP), which interfaced with the VICON system to record the raw ground truth data. For each target object, a CSV file was written containing the object position $(X,Y,Z)$ and orientation (pitch, roll, yaw) in VICON world coordinates together with the number of markers detected by the system, the VICON frame number and synchronization information.

\subsection{Data Synchronization}

A custom built synchronization circuit was connected via serial port to the data acquisition PC and used to trigger a camera flash near the start and end of each video recording. This allowed the two flash events to be logged within a designated synchronization field of each raw CSV file produced by the capture tool. The two video frames containing each flash were subsequently specified manually and these correspondences were used on the fly by the toolkit to line up in time each video frame with its nearest VICON frame.

\subsection{Camera Calibration}
Our camera calibration and pose estimation procedures both leverage OpenCV library implementations of~\cite{BouguetToolbox} and~\cite{zhang2000flexible} through its Python API to calculate the intrinsic and extrinsic camera parameters.

\begin{figure*}[!t]
\centering
\subfloat[]{\includegraphics[width=3.07in,trim={1cm 1.6cm 0.7cm 2.1cm},clip]{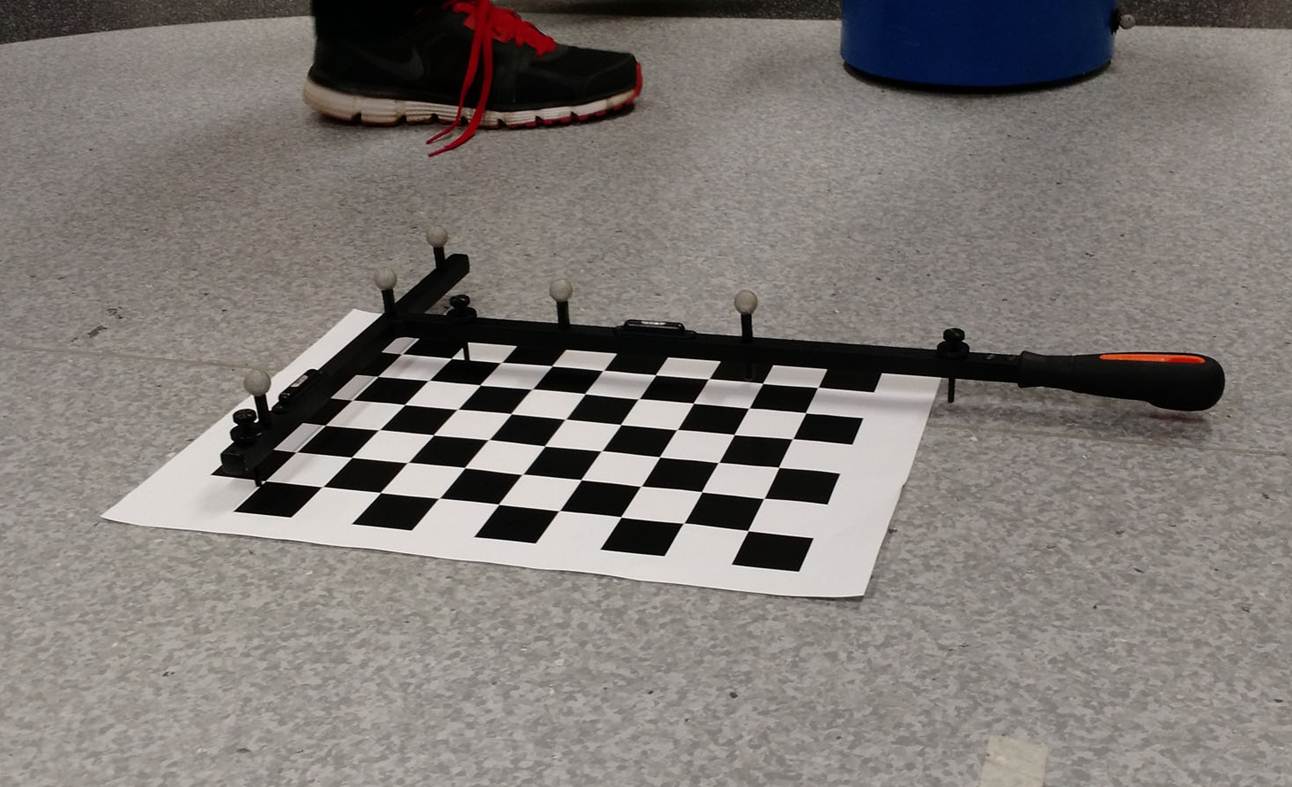}
\label{fig_first_case}}
\hfil
\subfloat[]{\includegraphics[width=1.5in,trim={1.35cm 0cm 0cm 0},clip]{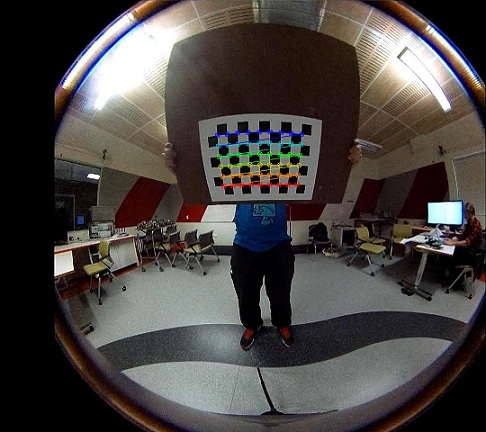}
\label{fig_second_case}}
\caption{ (a) Objects used for camera calibration: $9 \times 6$ chessboard pattern with 35.3 mm side squares and VICON calibration wand with five markers. (b) RICOH THETA m15 single lens view of the chessboard pattern, showing the image training points extracted by OpenCV automatic corner detection.}
\label{fig_calibration}
\end{figure*}

\subsubsection{Intrinsic Parameters}

A perspective projection model was computed separately for each lens using the camera calibration tool (CLB). This used the corner points extracted from multiple views of a chessboard pattern as input, and these images are provided as part of the data set, with one example shown in Figure~\ref{fig_calibration}. The parameters include the focal lengths ($f_x$ and $f_y$) and principal point coordinates ($c_x$ and $c_y$), which are in pixel units and serve as the elements of the \emph{camera matrix} $\textbf{A}$ in the following transform~\cite{OpenCVDocs}:

\begin{equation}
\begin{gathered}
s \; \textbf{m} = \textbf{A} [\textbf{R}|\textbf{t}] \textbf{M}\\
s \begin{bmatrix} u \\ v \\ 1 \end{bmatrix}
=     
\begin{bmatrix}
       f_x & 0 & c_x \\
       0 & f_y & c_y \\
       0 & 0 & 1
     \end{bmatrix}
\begin{bmatrix}
       r_{11} & r_{12} & r_{13} & t_1 \\
       r_{21} & r_{22} & r_{23} & t_2 \\
       r_{31} & r_{32} & r_{33} & t_3 
     \end{bmatrix}
\begin{bmatrix} X \\ Y \\ Z \\ 1 \end{bmatrix}
\end{gathered}
\label{withoutdist}
\end{equation}

Ignoring distortion, Eqn.~\ref{withoutdist} describes the projection of a corner position $\textbf{M}$ in world coordinates defined by the chessboard plane (such that $Z=0$) to its corresponding image point $\textbf{m}$ with pixel coordinates ($u,v$),  where $s$ is a scale factor. The rotation matrix $\textbf{R}$ and translation vector $\textbf{t}$ relate the (fixed) camera pose to the coordinate system defined by a given chessboard orientation.

To handle real cameras, the intrinsic parameters also include distortion coefficients, which are used to model the radial ($k_{1}, ..., k_{6}$) and tangential ($p_{1},p_{2}$) image distortion in Eqn.~\ref{withdist} below. In order to handle the strong radial distortion induced by the fisheye projection, we applied the OpenCV \emph{rational} model, which activates three additional radial coefficients: $k_{3}, k_{4}, k_{6}$. This provides a denominator term Eqn.~\ref{withdist}, which would otherwise be $1$ under default settings. Expanding out Eqn.~\ref{withoutdist} to an equivalent series of expressions, this distortion model is incorporated into the transform as follows~\cite{OpenCVDocs}: 

\begin{equation}
\begin{gathered}
\begin{pmatrix} x\\ y\\ z \end{pmatrix} = \textbf{R} \begin{pmatrix} X\\ Y\\ Z \end{pmatrix} + \textbf{t} \\
x' = x/z \\
y' = y/z \\ 
x'' = x' \frac{1 + k_1 r^2 + k_2 r^4 + k_3 r^6}{1 + k_4 r^2 + k_5 r^4 + k_6 r^6} \\
+ 2 p_1 x' y' + p_2(r^2 + 2 x'^2)
\\ y'' = y' \frac{1 + k_1 r^2 + k_2 r^4 + k_3 r^6}{1 + k_4 r^2 + k_5 r^4 + k_6 r^6} \\
+ p_1 (r^2 + 2 y'^2) + 2 p_2 x' y'
\\ \text{where} \quad r^2 = x'^2 + y'^2
\\ u = f_x \times x'' + c_x
\\ v = f_y \times y'' + c_y 
\end{gathered}
\label{withdist}
\end{equation}

\subsubsection{Extrinsic Parameters}
The camera pose estimation must be performed separately for each lens and for each of the four data capture sessions. The procedure relies on a set of 2D image and 3D object training point correspondences, where each point is the marker at the `T-junction' of the VICON calibration wand seen in Figure~\ref{fig_calibration}. Training sets for each lens were built by performing the data acquisition with the wand as the target to collect object points. A human operator then used the training tool (TRN) to annotate a number of image points through mouse clicks on the marker at different frames, as the wand moved throughout the room. While a minimum of $4$ training points are required for each lens, in excess of $20$ were used in practice. These were output by the training tool in XML file format and are provided as part of the data set.

Given the appropriate training set and intrinsic parameters, the camera pose is estimated on the fly for each lens by the mapping tool (MAP). The training point correspondences and fixed intrinsic parameters are passed to the OpenCV function \emph{solvePnP}, where the camera pose is first estimated by the Direct Linear Transform method and then optimized through an iterative procedure based on the Levenberg-Marquardt algorithm. This finds the rotation matrix $\textbf{R}$ and translation vector $\textbf{t}$ necessary to describe the transformation from the VICON world coordinate system to a camera world coordinate system.

\subsection{Mapping Ground Truth Positions}
Once the camera was calibrated and a pose estimation training set had been built, the VICON coordinates of new target objects were mapped automatically in order to generate ground truth data for every video. 

\subsubsection{Switching between the lenses}
In order to perform the mapping, the appropriate lens must first be selected. To this end, the mapping tool applies extrinsic parameters of each lens to each VICON position $(X,Y,Z)$ in order to convert the object position to camera world coordinates for that particular lens:
\begin{equation}
\begin{pmatrix} x\\ y\\ z \end{pmatrix} = \textbf{R} \begin{pmatrix} X\\ Y\\ Z \end{pmatrix} + \textbf{t} \\
\end{equation}
By converting this into spherical coordinates:
\begin{equation}
\begin{gathered}
 r = \sqrt{x^2 + y^2 + z^2} \\
 \theta = \arccos\frac{z}{\sqrt{x^2 + y^2 + z^2}} \\
 \phi = \arctan \frac{y}{x}
\end{gathered}
\end{equation}
the zenith angle $\theta$ is used to determine in which of the hemispheres the object is located, and hence which set of parameters should be used to perform the mapping. 

\subsubsection{Mapping 3D positions to 2D centroids}

\begin{figure*}[!t]
\centering
\subfloat[]{\includegraphics[width=2.2in]{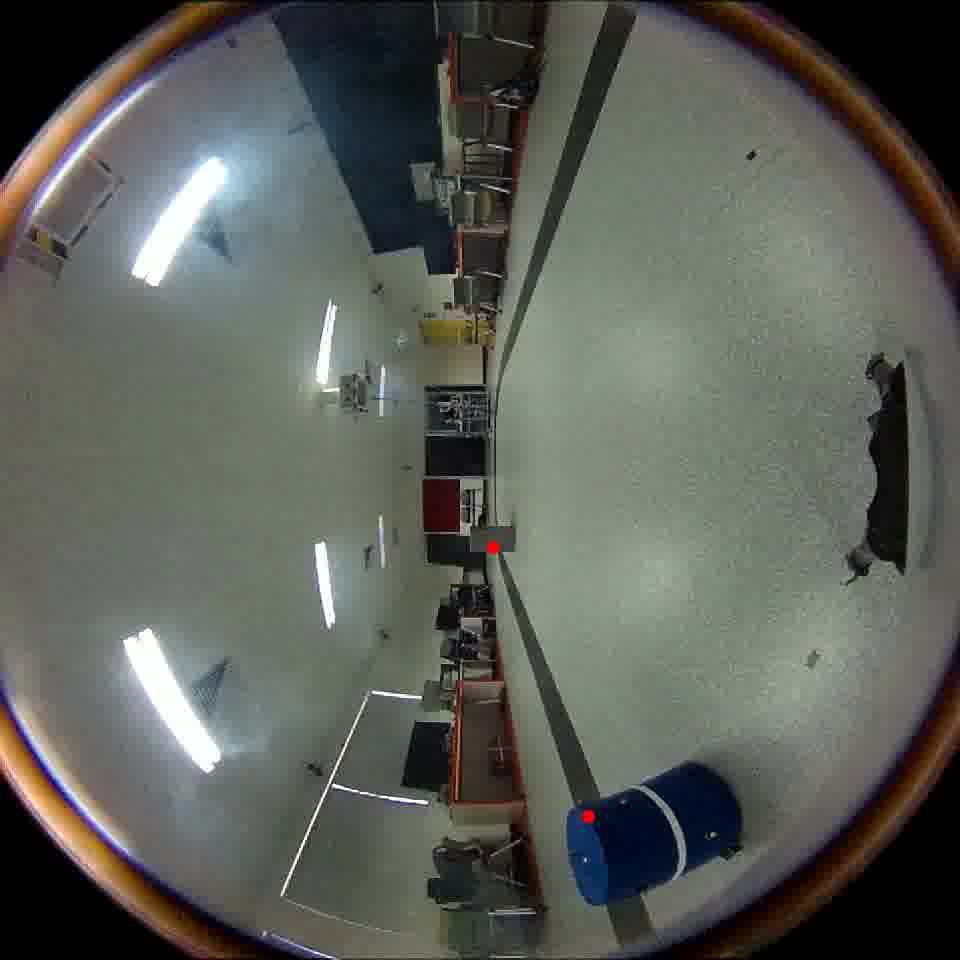}
\label{fig_backside}}
\hfil
\subfloat[]{\includegraphics[width=2.2in]{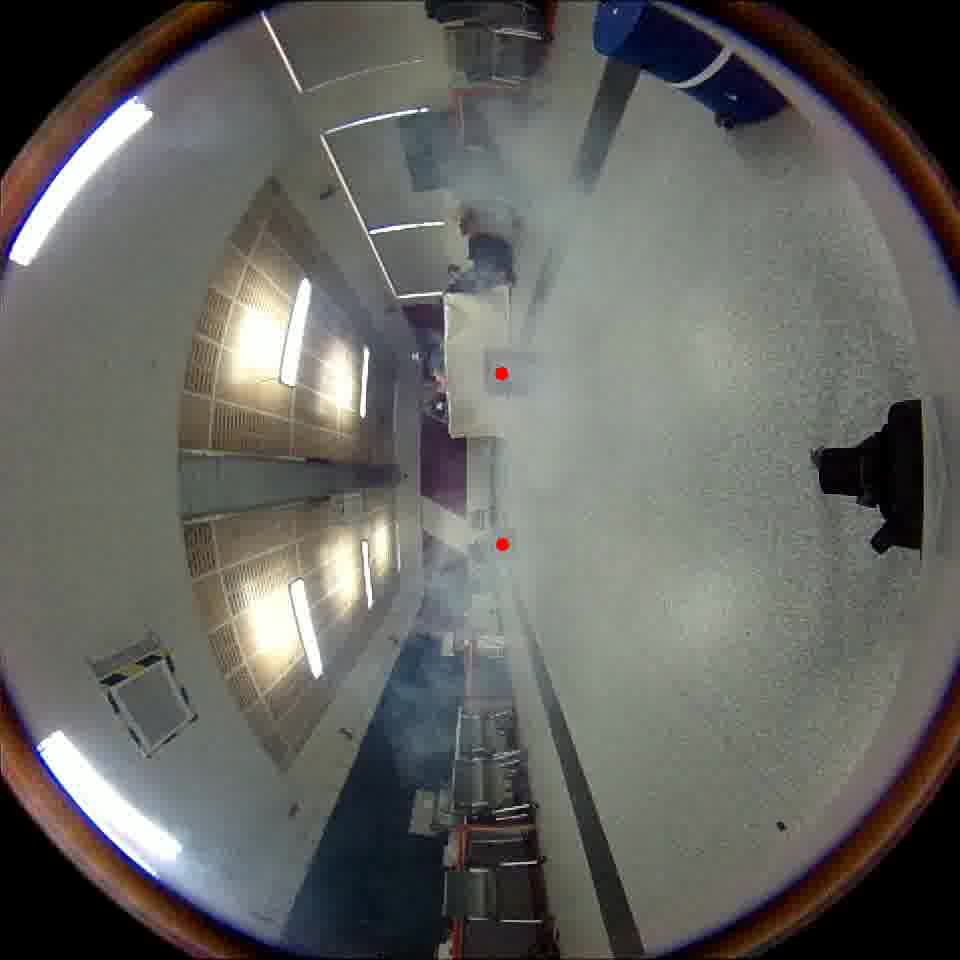}
\label{fig_buttonside}}
\hfil
\subfloat[]{\includegraphics[width=2.55in,trim={3.7cm 0cm 3.6cm 0},clip]{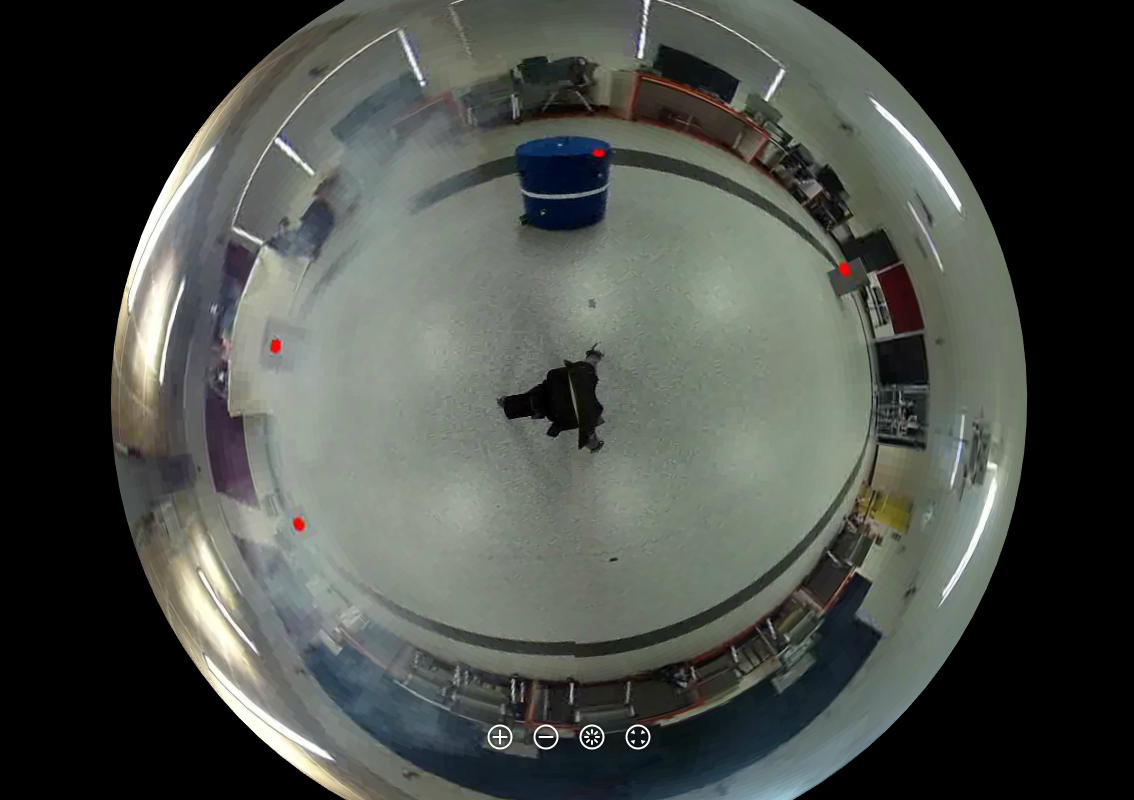}}
\label{fig_app}
\caption{Mapped ground truth annotations (red points) in video `Set 33' of session 4 displayed in the (a) left and (b) right side images of a video frame, which can be stitched into an equirectangular image and displayed in (c) using the RICOH THETA spherical viewer~\cite{RICOHTHETAapp}. Of the four target objects (EE1, EE3, EE4, EE5) in this frame, two are obscured by fog from a fog machine.}
\label{fig_fog}
\end{figure*}

The projection into the image of object positions measured by the VICON system was performed using the mapping tool according to Eqn.~\ref{withdist}, where the extrinsic parameters \textbf{R} and \textbf{t} were computed from the wand training points. Table~\ref{tab:unisa} presents a overview of the the $43$ mapped videos contained in the UniSA omnidirectional data set. Figure~\ref{fig_fog} illustrates some sample results in two fisheye images. Furthermore, these can be stitched and converted into an equirectangular image by using the RICOH THETA Spherical Viewer~\cite{RICOHTHETAapp}, which displays it on the sphere.

\begin{table*}[h]
  \centering
    \begin{tabular}{ | l | l | l | l | l | l | p{7cm} |}
    \hline
     Name & Video & Frames& Start & End& Targets & Scenario Notes \\ \hline
     Session 1 Set 1 & R0010216 & 325 & 36 & 362 & EE2 & Single target, still \\ \hline
     Session 1 Set 2 & R0010218 & 324 & 37 & 362 & EE2 & Single target, moving from off-screen\\ \hline
     Session 1 Set 3 & R0010219 & 324 & 31 & 356 & EE2 & Single target, moving on-screen, across two lens\\ \hline
     Session 1 Set 4 & R0010220 & 341 & 14 & 356 & EE2 & Single target, moving on-screen, single lens\\ \hline
     Session 1 Set 5 & R0010221 & 324 & 37 & 362 & EE2 & Single target, full occlusion\\ \hline
     Session 1 Set 6 & R0010222 & 334 & 27 & 354 & EE2 & Single target, partial occlusion\\ \hline
     Session 1 Set 7 & R0010224 & 324 & 30 & 355 & EE2 & Single target, lighting variation\\ \hline
     Session 1 Set 8 & R0010226 & 335 & 34 & 370 & EE2, EE5 & 2 targets, separate lens\\ \hline
     Session 1 Set 9 & R0010227 & 324 & 36 & 361 & EE2, EE5 & 2 targets, same lens, no occlusion\\ \hline
     Session 1 Set 10 & R0010228 & 333 & 28 & 362 & EE2, EE5 & 2 targets, same lens, partial occlusion\\ \hline
     Session 1 Set 11 & R0010229 & 326 & 35 & 362 & EE2, EE5 & 2 targets, same lens, multi occlusion\\ \hline
     Session 1 Set 12 & R0010230 & 329 & 36 & 366 & EE2, EE4, EE5 & 3 targets, 1 stationary, multi occlusion with lighting variations\\ \hline
     Session 1 Set 13 & R0010232 & 342 & 23 & 348 & EE2, EE3, EE4, EE5 & 4 targets, 2 stationary, separate lenses, partial occlusion\\ \hline
     Session 1 Set 14 & R0010233 & 325 & 34 & 360 & EE2, EE3, EE4, EE5 & 4 targets, 2 stationary, separate lenses, full occlusion\\ \hline
     Session 2 Set 15 & R0010238 & 326 & 32 & 359 & EE1, EE2, EE4, EE5 & 4 targets moving in sectors, no crossing lens \& occlusions\\ \hline
     Session 3 Set 16 & R0010242 & 337 & 42 & 380 & EE1, EE2, EE3, EE4, EE5, EE6 & 6 targets, no occlusions, 1 cross lens\\ \hline
     Session 3 Set 17 & R0010243 & 341 & 34 & 376 & EE1, EE2, EE3, EE4, EE5, EE6 & 6 targets, varied occlusions, 1 cross lens\\ \hline
     Session 3 Set 18 & R0010244 & 349 & 41 & 391 & EE1, EE2, EE3, EE4, EE5, EE6 & 5 moving targets, 1 stationary target, rotation, occlusions\\ \hline
     Session 3 Set 19 & R0010245 & 1260 & 26 & 1287 & EE1, EE2, EE3, EE4, EE5, EE6 & 6 targets, long set, loop around blue, occlusions\\ \hline
     Session 3 Set 20 & R0010246 & 1349 & 38 & 1388 & EE1, EE2, EE3, EE4, EE5, EE6 & 6 targets, long set, lots of clutter with chairs\\ \hline
     Session 3 Set 21 & R0010247 & &  &  &  & Reference frame, no targets\\ \hline
     Session 3 Set 22 & R0010248 & 1256 & 40 & 1297 & EE1, EE2, EE4, EE5, EE6 & 5 targets, long set, reveal from off-scene\\ \hline
     Session 3 Set 23 & R0010249 & 1257 & 43 & 1301 & EE1, EE2, EE4, EE5, EE6 & 5 targets, long set, chasing humanoid object, occlusions\\ \hline
     Session 3 Set 24 & R0010250 & 324 & 48 & 373 & EE1, EE2, EE4, EE5, EE6 & 5 targets, moving before, stop mid-scene with clutter\\ \hline
     Session 3 Set 25 & R0010251 & 324 & 36 & 361 & EE1, EE2, EE4, EE5, EE6 & 5 targets, spinning on spot\\ \hline
     Session 3 Set 26 & R0010252 & 319 & 41 & 367 & EE1, EE2, EE3, EE4, EE5, EE6 & 5 targets, spinning counter\\ \hline
     Session 3 Set 27 & R0010253 & 324 & 46 & 371 & EE1, EE2, EE4, EE5, EE6 & 5 targets, with drivers as clutter\\ \hline
     Session 3 Set 28 & R0010254 & 324 & 41 & 366 & EE1, EE2, EE4, EE5, EE6 & 5 targets, lighting changes on backside\\ \hline
     Session 3 Set 29 & R0010255 & 328 & 35 & 364 & EE1, EE2, EE4, EE5, EE6 & 5 targets, lighting changes on buttonside\\ \hline
     Session 3 Set 30 & R0010256 & 327 & 42 & 370 & EE1, EE2, EE4, EE5, EE6 & 5 targets, lighting changes on buttonside\\ \hline
     Session 4 Set 31 & R0010259 &  &  &  &  & Reference frame, no targets\\ \hline
     Session 4 Set 32 & R0010261 & 335 & 35 & 371 & EE1, EE3, EE4, EE5 & 4 targets, a little fog\\ \hline
     Session 4 Set 33 & R0010262 & 324 & 40 & 365 & EE1, EE3, EE4, EE5 & 4 targets, a lot of fog\\ \hline
     Session 4 Set 34 & R0010263 & 338 & 41 & 365 & EE1, EE3, EE4, EE5 & 4 targets, misty fog with total lighting changes\\ \hline
     Session 4 Set 35 & R0010264 & 332 & 44 & 377 & EE1, EE3, EE4, EE5 & 4 targets, fog pointed at camera, single side lighting changes\\ \hline
     Session 4 Set 36 & R0010265 & 346 & 35 & 382 & EE1, EE3, EE4, EE5 & 4 targets, no fog, complete dark start\\ \hline
     Session 4 Set 37 & R0010266 & 339 & 42 & 382 & EE1, EE3, EE4, EE5 & 4 targets, fog with spinning\\ \hline
     Session 4 Set 38 & R0010267 & 325 & 42 & 368 & EE1, EE3, EE4, EE5 & 4 targets, fog with anti-clockwise rotation\\ \hline
     Session 4 Set 39 & R0010268 & 328 & 35 & 372 & EE1, EE3, EE4, EE5 & 4 targets, no fog, all robots occlude behind a single robot\\ \hline
     Session 4 Set 40 & R0010269 & 328 & 46 & 375 & EE1, EE3, EE4, EE5 & 4 targets, no fog, all robots occlude behind a single robot\\ \hline
     Session 4 Set 41 & R0010272 & 325 & 34 & 360 & EE1, EE3, EE4, EE5 & 4 targets, random stuff\\ \hline
     Session 4 Set 42 & R0010273 & 324 & 37 & 362 & EE1, EE3, EE4, EE5 & 3 moving targets, 1 stationary, occlusion, bumps\\ \hline
     Session 4 Set 43 & R0010274 & 324 & 39 & 364 & EE1, EE3, EE4, EE5 & 3 moving targets, 1 stationary, occlusion bumps\\ \hline
    \end{tabular}
    \caption{\label{tab:unisa}Overview of the UniSA omnidirectional data set}
\end{table*}

Although the annotations tend to look reasonable by eye, mapping errors do occur in some cases. Figure~\ref{fig_backside} shows one such case, where the annotation of the nearest object (EE1) is not well centered on the object. This is likely due to poor mapping near the edge of the lens field of view, where the distortion given by the fisheye projection is strong and not well modeled (see below). In section~\ref{conclusion} we outline a future direction for improving our current calibration method, which may overcome this type of problem. Here on the other hand we seek to quantify the systematic error on the ground truth annotations.

\subsubsection{Expected Error}
\begin{figure}[!t]
\centering
\includegraphics[width=3.2in,trim={5.9cm 0.4cm 4.0cm 6.0cm},clip]{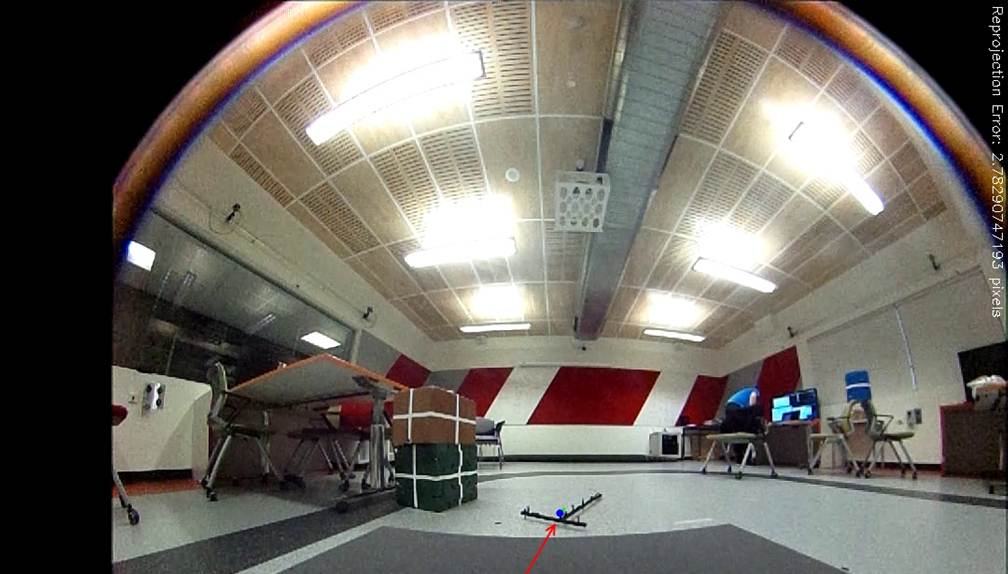}
\caption{The red arrow indicates a mapped image point (green) and the corresponding training point (blue) on the calibration wand marker. The green point has been obtained by projecting the measured world coordinates of the marker using the mapping tool (MAP). The training point (blue) was labeled by a human using the training tool (TRN) and contributed to the determination of the extrinsic parameters used in the mapping. The Euclidean distance between the two points measures the re-projection error given by the mapping, which in this example is $2.78$ pixels.}
\label{fig_wandreprojection}
\end{figure}

\begin{figure*}[!t]
\centering
\subfloat[]{\includegraphics[width=2.23in,trim={1.1cm 0cm 2.2cm 0.6cm},clip]{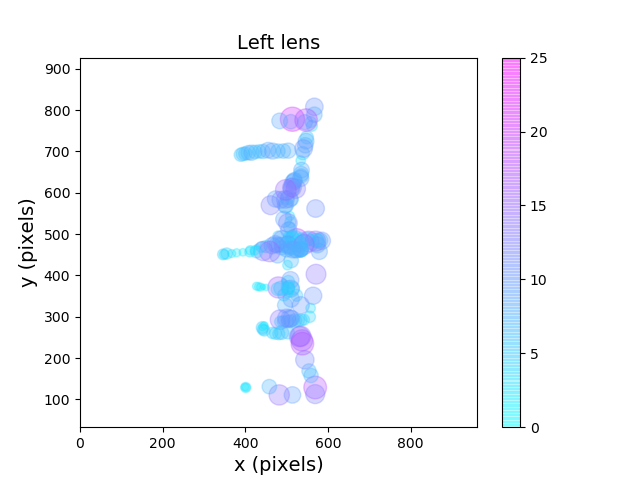}}
\hfil
\subfloat[]{\includegraphics[width=2.23in,trim={1.1cm 0cm 2.2cm 0.6cm},clip]{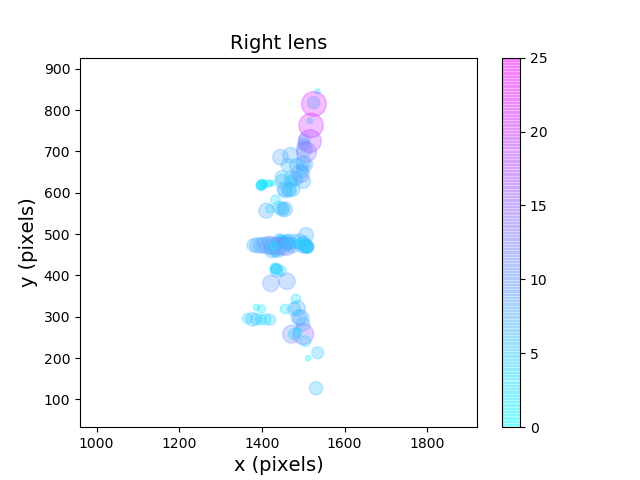}}
\caption{Wand re-projection errors in all sessions shown at their image training points in the (a) left and (b) right lens images, where the axes are oriented as those in Figures~\ref{fig_backside} and~\ref{fig_buttonside}, respectively. The re-projection error is indicated by the marker color in units of pixels, while the marker size also scales linearly with this value, such that larger markers indicate larger error.}
\label{fig_imageerr}
\end{figure*}

In order to estimate the systematic error on mapped centroids, the \emph{Compare Trainer} utility tool compared re-projected VICON calibration wand marker points with their corresponding image training points. Using the wand for this is necessary because it is known a priori precisely to which 3D position (marker) the measured VICON coordinates belong. Figure~\ref{fig_wandreprojection} provides an example of one such comparison, which captures errors induced by imperfections in both the camera model (including distortion) and the pose estimation. The Euclidean distances between the training and re-projection points were measured using all training examples for each of the two lenses and every data capture session. Their spatial distribution across each image, shown in Figure~\ref{fig_imageerr}, suggests that mapped points near the edge of the field of view are subject to some of the largest systematic errors, although examples of good mappings are also found in the outer regions. 
Tables~\ref{tab:session1},~\ref{tab:session2},~\ref{tab:session3} and ~\ref{tab:session4} summarize the re-projection error in terms of the sample \emph{mean} and \emph{standard deviation} $\sigma$, while the number of training points used is also listed.

\begin{table}[H]
\centering
\begin{tabular}{l| r| r | r}
 & \emph{mean (pixels)} & \emph{$\sigma$ (pixels)} &  Points \\\hline
\emph{Left lens} & 8.22 & 3.79 & 30 \\
\emph{Right lens} & 5.53 & 3.23 & 21
\end{tabular}
\caption{\label{tab:session1} Re-projection error in Session 1.}
\end{table}

\begin{table}[H]
\centering
\begin{tabular}{l| r| r | r}
 & \emph{mean (pixels)} & \emph{$\sigma$ (pixels)} & Points \\\hline
\emph{Left lens} & 7.23 & 4.06 & 72 \\
\emph{Right lens} & 5.80 &  4.21 & 56
\end{tabular}
\caption{\label{tab:session2} Re-projection error in Session 2.}
\end{table}

\begin{table}[H]
\centering
\begin{tabular}{l| r| r | r}
 & \emph{mean (pixels)} & \emph{$\sigma$ (pixels)} & Points \\\hline
\emph{Left lens} & 6.73 & 3.21 & 48 \\
\emph{Right lens} &  5.88 &  3.41 & 30
\end{tabular}
\caption{\label{tab:session3} Re-projection error in Session 3.}
\end{table}

\begin{table}[H]
\centering
\begin{tabular}{l| r| r | r}
 & \emph{mean} (pixels)& \emph{$\sigma$ (pixels)} & Points \\\hline
\emph{Left lens} &  6.82 & 4.31 & 36 \\
\emph{Right lens} & 5.17 & 2.52 & 22
\end{tabular}
\caption{\label{tab:session4} Re-projection error in Session 4.}
\end{table}

\section{Performance Evaluation Framework} \label{performance}
This section outlines the way in which the performance of visual tracking systems or human annotators may be compared and evaluated against the automatic ground truth annotations produced for the UniSA omnidirectional data set.

\subsection{Wizard GUI}
A graphical user interface (GUI) called Wizard
exists to guide users through the process summarized by Figure~\ref{fig_dataflow}. Rather than running tools from the command line, users can opt to replicate the work flows described in previous section using the Wizard, starting with the camera calibration and/or the trainer tools, or alternatively, by using the mapped ground truths provided. Furthermore, Wizard sessions save metadata in an XML header file, which maintains a record of input files and allows a zipped data set to be re-loaded.

\subsection{Ground Truth Data Format}

If re-calibration and re-mapping are not required, the user may simply compare (CMP) our ground truth data, which are the mapped image annotations, against results from their vision system. This requires both inputs to the comparison tool to have a common format, which is described below.

\begin{lstlisting}[basicstyle=\scriptsize]
<dataset>
  <frameInformation>
    <frame number="1" />
    <object name="EE1" lens="Backside" id="01">
        <boxinfo y="488" x="499" width="63" height="74"/>
        <centroid y="525" x="530"/>
        <visibility visible="5" visibleMax="5"/>
    </object>
    <object name="EE2" lens="Buttonside" id="02">
        <boxinfo y="406" x="1465" width="83" height="104"/>
        <centroid y="458" x="1506"/>
        <visibility visible="5" visibleMax="5"/>
    </object>
  </frameInformation>
</dataset>
\end{lstlisting}

The example above illustrates the ground truth annotation format. Each \emph{object name} is associated with a unique \emph{ID}. The choice of \emph{lens} is defined by which half of the $1920\times1080$ pixel frame the \emph{centroid} is located: \emph{Backside}/\emph{Buttonside} refers to the left/right image in the frame. The \emph{visbility} tag indicates the quality of the VICON tracking and this is used by the comparison tool to decide whether ground truth data is available for that object and at that frame. A \emph{boxinfo} attribute is provided as a place holder and currently has rather arbitrary values that prescribe the corner and dimension of a rectangular bounding box centred on the \emph{centroid}. The reason behind this is that only the projected position (and not the projected shape) is available. In future releases we intend to refine the bounding boxes and automatically scale them with the distance to the object.

\subsection{Evaluation of vision systems}
The evaluation tool (CMP) provides a simple frame-by-frame calculation of the the Euclidean distance between each ground truth annotation and the position of that object output by vision system. Here we provide a semi-automated annotation tool (ANT) based on~\cite{aripinotator} as a demonstrator for a vision system that outputs its results in the required format. Comparisons are made for ground truth and system objects with the same name and on the same lens.  One of the requirements for the vision system is therefore to assign the correct ID to the each object, in every frame and multi-object trackers can perform this task. The comparison tool output includes frame-by-frame display of the Euclidean error, which can be toggled between each ground truth object present in the video.

\section{Conclusion}  \label{conclusion}

This paper has presented a novel approach to automatic ground truth annotation for multi-object visual tracking. To our best knowledge, the idea of automatically mapping 3D ground truth positions into the image has not been implemented without input from human annotators in public benchmarks for visual tracking. We use our approach to create the UniSA omnidirectional data set, which provides the research community with $43$ spherical annotated videos of moving and stationary targets under a variety of challenging scenarios. We also supply the raw world coordinate data and raw calibration data.

Based on the video and ground truth data provided, the UniSA omnidirectional data set may be used to train and evaluate algorithms for the object detection, tracking and fine-grained recognition of multiple objects. In the latter case, the omnidirectional vision system could, for example, be required to detect all target objects while ignoring all others (e.g. people and furniture) considered to be clutter, and then reconize a particular object type (e.g. only the elongated containers).

A key aspect of this data set is that the camera can be re-calibrated (e.g. with a refined camera model) and the ground truth annotations can be subsequently re-mapped. To this end, we plan to investigate the application of the spherical OcamCalib Toolbox for Matlab~\cite{scaramuzza2006flexible} to our data. An outstading issue is that unlike the toolbox by Bouguet~\cite{BouguetToolbox}, which is implemented in OpenCV to handle 3D calibration structures, OCamCalib only solves the extrinsic parameters of planar homographies, and so can not be directly applied to find our extrinsic parameters based on calibration wand points. Nevertheless, we expect that any future release of our data set will involve the application of its spherical camera model to handle the perspective projection of the dual fisheye lens camera.


\section*{Acknowledgment}

The authors would like to thank the University of South Australia School of Engineering for the use of the Mawson Lakes campus Mechatronics Laboratory.




\bibliography{IEEEabrv.bib,omnirefs.bib}{}
\bibliographystyle{IEEEtran}

%
%
%

%

\end{document}